\documentclass[conference]{IEEEtran}
\IEEEoverridecommandlockouts
\usepackage{cite}
\usepackage{amsmath,amssymb,amsfonts}
\usepackage{algorithmic}
\usepackage{graphicx}
\usepackage{textcomp}
\usepackage{xcolor}
\def\BibTeX{{\rm B\kern-.05em{\sc i\kern-.025em b}\kern-.08em
    T\kern-.1667em\lower.7ex\hbox{E}\kern-.125emX}}
\usepackage{multirow}
\usepackage{tikz}
\usepackage{pgfplots}
\pgfplotsset{compat=1.17}
\usepgfplotslibrary{groupplots}
\usepackage{lipsum}
\usepackage{arydshln}
\usepackage{hyperref}
\definecolor{mypurple}{rgb}{0.59, 0.44, 0.84}

\begin{document}
\title{Few-shot Decoding of Brain Activation Maps}

\author{\IEEEauthorblockN{Myriam~Bontonou, Giulia~Lioi, Nicolas~Farrugia, Vincent~Gripon}
\IEEEauthorblockA{\textit{Lab-STICC, IMT Atlantique} \\
Brest, France  \\
firstname.surname@imt-atlantique.fr}
}

\maketitle

\begin{abstract}
Few-shot learning addresses problems for which a limited number of training examples are available. So far, the field has been mostly driven by applications in computer vision. Here, we are interested in adapting recently introduced few-shot methods to solve problems dealing with neuroimaging data, a promising application field. To this end, we create a neuroimaging benchmark dataset for few-shot learning and compare multiple learning paradigms, including meta-learning, as well as various backbone networks. Our experiments show that few-shot methods are able to efficiently decode brain signals using few examples, which paves the way for a number of applications in clinical and cognitive neuroscience, such as identifying biomarkers from brain scans or understanding the generalization of brain representations across a wide range of cognitive tasks.
\end{abstract}


\section{Introduction}
\label{sec:intro}

Machine learning, in particular deep learning, has achieved unprecedented performance in a variety of tasks in the past few years. As a matter of fact, when training data is abundant, deep neural networks (DNNs) are often able to extract relevant hidden features to generalize decisions to new data. The need for a large amount of training data is easily explained by the fact that DNNs contain a very large number of trainable parameters, which are required to achieve the best performance. 

However, in some domains, data labeling or acquisition can be expensive, thus limiting the potential of deep learning-based methods. In particular, in the case of neuroscience, studies require a massive investment in material and human resources. In this article alone, we use neuroimaging data from a study where 13 people have spent several hours in MRI scanners. And yet, we are far from the amount of data usually required to train a DNN. In an effort to tackle this limitation, the field of few-shot learning has proposed a lot of methods in the past few years (c.f. Section~\ref{sec:related}). A vast majority of these works focus on the case of computer vision, with the double interest of the relative simplicity of the considered signals and the existence of standardized benchmarks.

Here, our primary aim is to stress the ability of recently introduced few-shot methods to solve tasks in the domain of neuroimaging. To this end we compare several few-shot methods on neuroimaging data coming from the collection of the IBC dataset (releases 1~\cite{IBC2018} and 2~\cite{pinho2020individual}) on NeuroVault~\cite{gorgolewski2015neurovault}. Our contribution is twofold:
\begin{itemize}
    \item we create and introduce a neuroimaging classification benchmark for few-shot learning,
    \item we adapt several vision-based few-shot methods and compare their performance.
\end{itemize}
The benchmark and our PyTorch code are released publicly on {\color{cyan} \small{\url{https://github.com/mbonto/fewshot_neuroimaging_classification}}}.

After carefully selecting the data from the IBC collection, the new benchmark ends up with 106 classes and about 20 examples per class. The classes are conditions linked to tasks performed by subjects during a fMRI scan (e.g. audio sentence, speech sound, face gender, reward, left hand...). The examples are elementary contrast maps derived from the fMRI scans. As these inputs are naturally supported on brain graphs, this benchmark can also be used to test graph-based methods. 

The structure of the paper is as follows. In Section~\ref{sec:related}, we introduce related work in the domains of signal processing for neuroimaging and few-shot classification. In Section~\ref{sec:methodology}, we present the methodology, including the creation of the neuroimaging benchmark for few-shot classification, together with the main backbone networks and methods we consider in this work. In Section~\ref{sec:experiments}, we conduct experiments aiming at: 1)~stressing the ability of few-shot methods to achieve a better accuracy than a direct classifier trained with a limited amount of data, and 2)~comparing the architectures of backbones in terms of signal processing.
Finally, Section~\ref{sec:conclusion} is a conclusion.

To the best of our knowledge, few-shot learning has never been applied to neuroimaging, a field where modern deep learning approaches have been applied to the analysis of large datasets only recently~\cite{schulz2020different, abrol2020hype, he2020meta}. This work makes an important step towards the application of deep learning to decode brain signals where generally few annotated examples are available. As it aims at fully exploiting the predictive power of DNNs for neuroimaging data, it may impact a number of applications (e.g. identifying biomarkers from brain scans, cognitive neuroscience\dots).

\section{Related work}
\label{sec:related}
\begin{figure*}[t]
\centering
\includegraphics[width=\textwidth, height=0.2\textwidth]{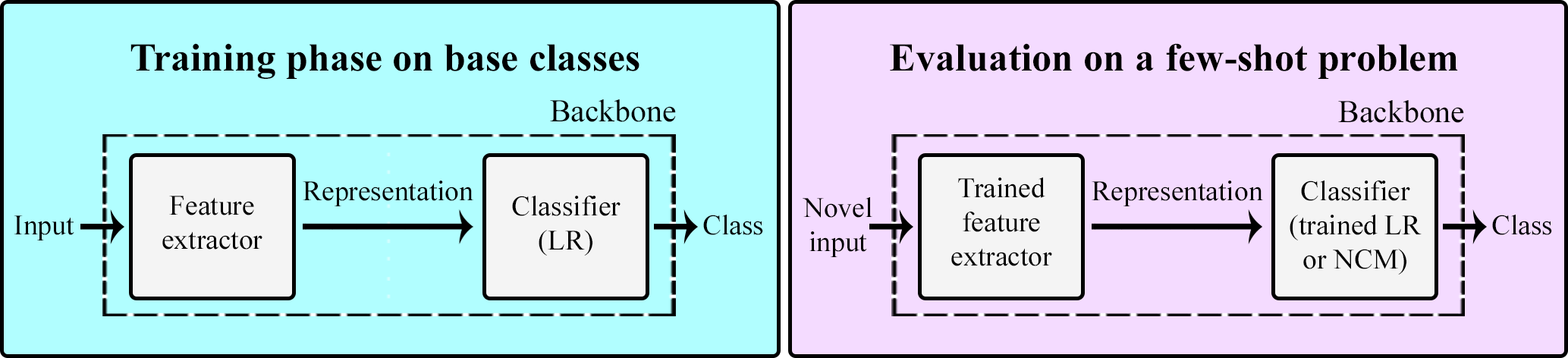}
\caption{\textbf{Overview of the few-shot learning approach.} During the training phase, the backbone learns rich representations from the abundant examples of base classes. During the evaluation phase, the backbone is adapted to new classes using only a few examples.}
\label{fig:FSL}
\end{figure*}
\begin{figure*}[t]
\centering
\includegraphics[width=\textwidth, height=0.23\textwidth]{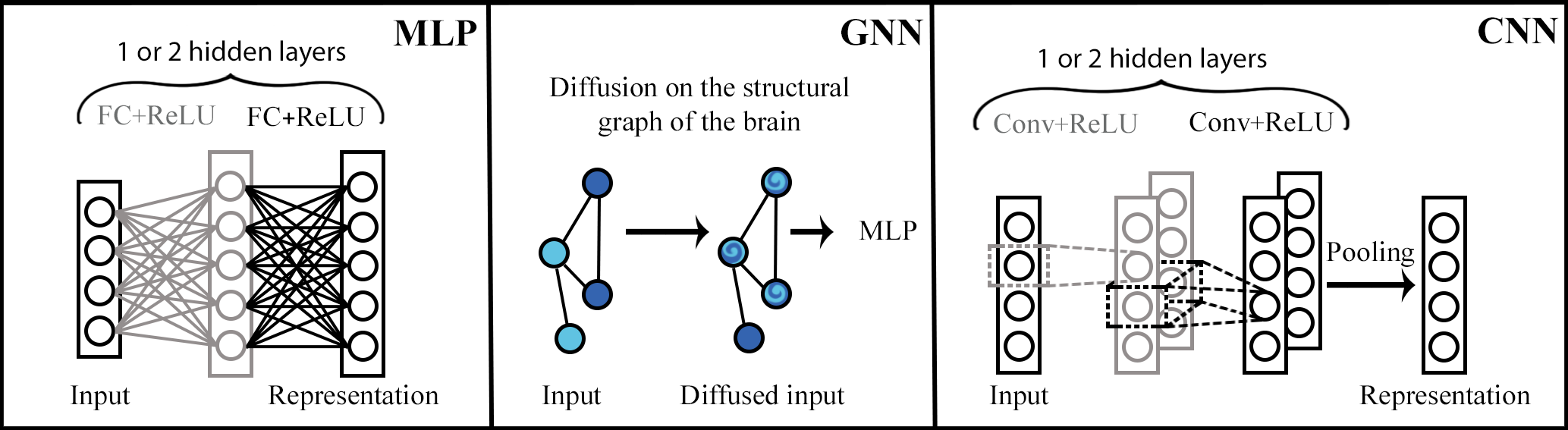}
\caption{\textbf{Architectures of the feature extractors.} FC stands for fully connected layer and Conv for convolutional layer with $1 \times 1$ kernels.}
\label{fig:Backbones}
\end{figure*}

Functional MRI (fMRI) measures brain activity changes caused by the deoxygenation of hemoglobin following neuronal activation (the so-called blood oxygen level dependent, or BOLD signal)~\cite{buxton2009introduction}, paving the way for human brain mapping.
While most classical approaches relied on simple linear statistical models of individual signals considered separately~\cite{friston1994statistical}, machine learning is increasingly being applied to decode brain activity~\cite{varoquaux2019predictive}, by performing supervised learning of multivariate activation patterns elicited by various tasks~\cite{kriegeskorte2006information,kriegeskorte2011pattern,haynes2006decoding}. As brain signals are naturally supported by the connectivity structure of the brain, recent studies suggest using graph neural networks (GNN) to decode brain activity by taking into account brain connectivity~\cite{ZHANG2021117847,Kazeminejad2019,Ktena2017}. In addition, a few recent studies have considered transfer learning scenarios; a large dataset is used to train a first classifier, and a smaller dataset exploits the representations learned by the first classifier in a fixed downstream task~\cite{zhang2020transferability,he2020meta}. Here, we consider the more specific setting of few-shot learning where the downstream task contains very few annotated elements.

The general idea of few-shot learning consists in training a DNN (called \emph{backbone} network) on a lot of data, and then, to transfer it as a prior to solve few-shot problems. The 3 main distinct strategies in the field are the following.

\emph{Representation-based} paradigm~\cite{wang2019simpleshot, chen2018closer, mangla2020charting, lichtenstein2020tafssl, hu2020leveraging, hu2020exploiting}: the first layers of the backbone network are seen as a feature extractor, and the last layer as a simple classifier (e.g. logistic regression). The strategy is to learn, on many classes, how to extract relevant hidden features. When a few-shot problem comes, the features of the few data examples are extracted with the feature extractor. A classifier, containing few parameters, is then trained on these features to recognize new classes.

\emph{Initialization-based} paradigm~\cite{finn2017model, antoniou2018train, nichol2018first} (also called meta-learning): the backbone network is trained on many few-shot problems. A problem consists in a set of a few training examples and a set of target samples we aim at classifying. In a first loop, the backbone network trains on the training examples in a few epochs. In a second loop, the initial parameters of the backbone network are updated to minimize its loss evaluated on the target samples. Thus, on a novel few-shot problem, training the backbone in a few epochs should lead to a good performance.

\emph{Hallucination-based} paradigm~\cite{zhang2019few, chen2019image}: the number of training examples is artificially augmented, either by deforming the original examples, or by combining them. 

In this work, we focus on the comparison between representation-based and initialization-based, as the metrics used to evaluate hallucination-based methods may be difficult to setup objectively for brain signals.

\section{Methodology}
\label{sec:methodology}

In this section, we present the standard few-shot classification setting on images. We introduce the few-shot methods we compare in the experiments. We also detail how the neuroimaging dataset is constructed and adapted to the few-shot setting.

\subsection{Few-shot setting}
In few-shot learning, we use three datasets. The first one, called \emph{base dataset}, contains a large number of labeled data samples corresponding to many classes. The second one, called \emph{validation dataset}, is made of distinct classes. It is meant to evaluate the ability of a DNN trained on the base dataset to generalize to new classes. The final dataset, called \emph{novel dataset}, contains a few examples from novel classes distinct from both the base and the validation datasets. The idea is to exploit the knowledge acquired while training a network on the base dataset in order to circumvent the lack of data in the novel dataset. The network trained on the base dataset is called the \emph{backbone} network.

In the literature, few-shot tasks are often defined in the novel dataset as follows: we consider $N$-way (number of classes), $K$-shot (number of labeled examples per class), $Q$-query (number of samples to label per class) tasks. The aim is to correctly classify the $NQ$ unlabeled samples. Two types of methods can be implemented: \emph{inductive few-shot} where the $NQ$ unlabeled samples are classified independently, and \emph{transductive few-shot} where their joint distribution can be taken into account while performing classification.

\subsection{Learning paradigms}
We propose to evaluate the efficiency of three methods: SimpleShot~\cite{wang2019simpleshot}, PT+MAP~\cite{hu2020leveraging} and MAML++~\cite{antoniou2018train}. SimpleShot and PT+MAP aim at learning good representations for decoding images. They follow the representation-based paradigm. A backbone network is trained to extract generic features enabling to recognize all base classes at once, with the hope that these features will help to classify novel classes.
As for MAML++, it aims at learning a good initialization so that the backbone network can easily adapt to the novel dataset with only a few epochs of training, and as such it follows the initialization-based paradigm. SimpleShot and MAML++ are inductive, whereas PT+MAP is transductive. The purpose of PT+MAP is to adjust the distribution of each class in the representation space to fit a Gaussian-like distribution. Further details are in the original papers~\cite{wang2019simpleshot, antoniou2018train, hu2020leveraging}.

\subsection{Backbone networks}
The backbones, made up of a feature extractor and a logistic regression (LR), are trained on the base dataset. In the following, we call \emph{representation} the set of features obtained at the output of the feature extractor. When a new few-shot task comes, the backbones are adapted to new classes using only a few training examples. In MAML++, the whole backbone is fine-tuned for a few epochs. In SimpleShot, a nearest-class mean classifier (NCM) is applied on the representations of the inputs without additional training. Similarly, in PT+MAP, only a new classifier is trained. See Fig.~\ref{fig:FSL} for an illustration.

As mentioned earlier, recent studies advocate the use of GNN to classify neuroimaging data. Here, we try to exploit this graph data structure in the scope of few-shot classification of brain signals. We therefore consider three backbone architectures, described in the following and in Fig.~\ref{fig:Backbones}.

We first use a multi-layer perceptron (MLP) where the input is seen as a simple vector describing the activity of the considered regions of interest. We vary 2 hyperparameters: the number of hidden layers and the number of features per hidden layer. We then consider a simplified GNN based on the simple graph convolution model~\cite{wu2019simplifying}, where the input is the vector described before. It is diffused once on a structural graph describing the interaction between regions of interest, and then it is handled by a MLP. We vary the same hyperparameters as for MLP. Finally, we use a 1x1 convolutional neural network (CNN), considering each region of interest independently. Before the classifier, the feature maps are averaged per input region. We vary the number of hidden layers and the number of feature maps per hidden layer. For all backbones, as the size of our inputs are small (360), we only consider shallow neural networks with one or two hidden layers. We vary the number of feature (maps) between 64 and 1024. More details about the architectures and the graph are given on our \href{https://github.com/mbonto/fewshot_neuroimaging_classification}{GitHub}.

\subsection{Neuroimaging benchmark for few-shot learning}
The benchmark dataset contains 106 classes which are randomly split into 64 base classes, 21 validation classes and 21 novel classes. Each class has between 21 and 78 samples (median: 33) of size $105 \times 127 \times 105$. Here, we consider a summarized version of the samples into 360 regions of interest (ROI). The proposed few-shot-ready dataset is publicly available on the \href{https://github.com/mbonto/fewshot_neuroimaging_classification}{GitHub} of the project.
In the following, we give additional details explaining the use of the Individual Brain Charting (IBC) dataset~\cite{IBC2018}, the meaning of the classes and the content of the samples.

In the IBC dataset (release 1 and 2), 13 subjects perform 25 tasks covering many aspects of cognition, such as motor movements, emotions, language tasks, working memory, video watching or social tasks. Brain activity is measured simultaneously using functional MRI. Each task enables to study several conditions associated with a wide range of cognitive states. E.g. during the task called hcp\_motor, the studied conditions are movement of the left hand, right hand, left foot, right foot and tongue. In total, we included 106 conditions as classes in our benchmark. The full list is available on our GitHub. This variety of experiments makes IBC an interesting benchmark for few-shot learning in decoding cognitive states, as it enables to test whether the variability in a set of cognitive states can be exploited to decode new cognitive states with few examples.

In this benchmark, the data samples are not time-varying 3D measurements of brain activity. Instead, we use the \emph{contrast maps} available on NeuroVault~\cite{gorgolewski2015neurovault} (collection 6618). A contrast map is a statistical map of the effect of a condition on the brain activity of a subject. It is derived from a general linear model (GLM). Broadly speaking, the scans of each participant are aligned to each other and normalized into the standard MNI152 space. Then, they are analyzed using the GLM to capture variations in BOLD signal coupled to neuronal activity pertaining to task performance. This results in a set of weights across the whole brain, therefore considered as a contrast map of the effect of the condition on brain activity with respect to a baseline state. We have at least 21 samples of such maps, corresponding to multiple subjects and repetitions of the experiments.

Finally, in this work, we consider a summarized version of these contrast maps in regions of interest (ROIs). We perform a parcellation of these maps using the 360 ROI of the Glasser atlas~\cite{Glasser2016}, a recently proposed atlas of the human brain that integrates multimodal measures of the brain. For the GNN, we use a thresholded structural graph (1\% highest connection weights) defined on the same ROI, previously introduced in~\cite{preti2019decoupling} and used in~\cite{brahim2020graph}.

\section{Experiments}
\label{sec:experiments}

\begin{table*}[t]
\centering
\caption{Average accuracy and $95\%$ confidence interval over $10000$ $5$-way tasks from the novel dataset.}
\label{tab:best}
\vspace{0.25cm}
\small
\begin{tabular}{|c|c|c:c:c|c:c:c|}
\hline
\multirow{2}*{\shortstack{Method \\ \small{(setting)}}}
 & \multirow{2}*{\shortstack{\small{Archi}}}
 & \multicolumn{3}{c|}{Best backbones} 
 & \multicolumn{3}{c|}{Backbones with representations of size 360} \\
 &  & \multicolumn{1}{c}{\small{layers / features}} & \multicolumn{1}{c}{5-shot} & \multicolumn{1}{c|}{1-shot}
 & \multicolumn{1}{c}{\small{layers / features}} & \multicolumn{1}{c}{5-shot} & \multicolumn{1}{c|}{1-shot} \\
\hline
\small{Baseline} &- & - & $70.56 \pm 0.21$ & $57.26 \pm 0.20$
& - & $70.56 \pm 0.21$ & $57.26 \pm 0.20$\\
\hline
\multirow{3}*{\shortstack{\small{SimpleShot} \\ \small{(induct.)}~\cite{wang2019simpleshot}}} 
& \small{MLP} & $2 / 360$ & $\boldsymbol{86.00 \pm 0.16}$ & $\boldsymbol{72.54 \pm 0.20}$ 
& $2 / 360$ & $\boldsymbol{86.00 \pm 0.16}$ & $\boldsymbol{72.54 \pm 0.20}$\\
& \small{GNN} & $2 / 1024$ & $85.14 \pm 0.16$ & $71.96 \pm 0.21$
& $2 / 360$ & $84.96 \pm 0.16$ & $71.67 \pm 0.20$\\
& \small{CNN} & $2 / 64$ & $74.26 \pm 0.20$ & $59.98 \pm 0.20$
& $2 / 64$ & $74.26 \pm 0.20$ & $59.98 \pm 0.20$ \\
\hline
\multirow{3}*{\shortstack{\small{PT+MAP} \\ \small{(trans.)}~\cite{hu2020leveraging}}} 
& \small{MLP} & $2 / 360$ & $\boldsymbol{88.76 \pm 0.17}$ & $\boldsymbol{84.34 \pm 0.25}$
& $2 / 360$ & $\boldsymbol{88.76 \pm 0.17}$ & $\boldsymbol{84.34 \pm 0.25}$\\
& \small{GNN} & $2 / 1024$ & $87.86 \pm 0.18$ & $83.19 \pm 0.25$
& $2 / 360$ & $87.02 \pm 0.18$ & $82.39 \pm 0.25$\\
& \small{CNN} & $2 / 64$ & $74.82 \pm 0.20$ & $63.71 \pm 0.26$
& $2 / 64$ & $74.82 \pm 0.20$ & $63.71 \pm 0.26$\\
\hline
\multirow{3}*{\shortstack{\small{MAML++} \\ \small{(induct.)}~\cite{antoniou2018train}}} 
& \small{MLP} & $1 / 360$ & $82.31 \pm 0.18$ & $67.64 \pm 0.22$
& $1 / 360$ & $82.31 \pm 0.18$ & $67.64 \pm 0.22$\\
& \small{GNN} & $1 / 128$ & $80.87 \pm 0.18$ & $67.71 \pm 0.22$
& $1 / 360$ & $ 81.06 \pm 0.17$ & $66.47 \pm 0.22$\\
& \small{CNN} & $2 / 360$ & $76.80 \pm 0.22$ & $64.68 \pm 0.21$
& $2 / 360$ & $76.80 \pm 0.22$ & $64.68 \pm 0.21$\\
\hline
\end{tabular}
\end{table*}

Following the trend in few-shot learning, we consider the two most common case studies: $5$-way $5$-shot $15$-query and $5$-way $1$-shot $15$-query tasks. In practice, other settings could be studied depending on the application at stake.
The tasks are synthesized by randomly sampling classes and samples in these classes. To obtain statistically robust results on the novel dataset, we average the accuracy over $10000$ tasks and we report the confidence interval at $95\%$. As several hyperparameters have to be defined, we select them using the validation dataset. The ones obtaining the best accuracy on $500$ $5$-way $5$-shot tasks generated from this set are kept.
As stated in Section~\ref{sec:methodology}, the backbone networks are trained on the base dataset either to recognize all base classes at once (SimpleShot) or to directly classify few-shot problems (MAML++). When relevant, we use PyTorch's weighted random sampler to handle class imbalance.

\subsection{Comparison of few-shot learning paradigms}
We first compare the learning paradigms, while varying the backbone networks and their hyperparameters (number of layers, number of features per layer (MLP and GNN), number of feature maps (CNN)). The results are presented in Table~\ref{tab:best} (best backbones). We also compare the paradigms with a NCM directly applied on the raw data (baseline). Our first observation is that, in the inductive setting, SimpleShot outperforms MAML++. Interestingly, this is similar to experiments with standardized vision benchmarks in the field~\cite{wang2019simpleshot}. All considered paradigms allow significant increase in accuracy when compared to using the raw features, showing the ability of few-shot vision methods to perform well in the considered case of neuroimaging data. Finally, the best results are obtained using the transductive PT+MAP method, even though the gain is less significant when considering 5-shot tasks. As PT+MAP uses additional information from the query samples, it is not surprising.

\subsection{Comparison of backbones}
\begin{figure}[t]
  \centering








    
\pgfkeys{/pgf/number format/.cd,1000 sep={}}

\begin{tikzpicture}
\begin{groupplot}[group style={group size=2 by 1, horizontal sep=1cm, vertical sep=0cm}]

    \nextgroupplot[
    width=6cm, 
    height=6cm,
    ylabel near ticks, 
    xlabel near ticks,
	yticklabel style={font=\small}, 
	xticklabel style={font=\small, rotate=45},
    xlabel style={font=\small, align=center, at={(0.55,-0.2)}}, 
    legend style={at={(0.95,0.05)}, anchor=south east}, 
    ylabel style={font=\small, align=center}, 
    ylabel=Accuracy (\%), xlabel=Size of the representation, 
    xmode=log, 
    log ticks with fixed point, 
    scaled ticks=false, 
    xtick=data, 
    ymin=65, 
    ymax=80,
    xmin=50,
    xmax=1300]

    \addplot[color=blue, mark=triangle] table [x=Size, y=MLP-SS, col sep=comma] {csv/output.csv};
    \addplot[color=blue, mark=o] table [x=Size, y=GNN-SS, col sep=comma] {csv/output.csv};
    \addplot[color=blue, mark=square] table [x=Size, y=Conv1d-SS, col sep=comma] {csv/output.csv};
    \addplot[color=orange, mark=triangle] table [x=Size, y=MLP-MAML, col sep=comma] {csv/output.csv};
    \addplot[color=orange, mark=o] table [x=Size, y=GNN-MAML, col sep=comma] {csv/output.csv};
    \addplot[color=orange, mark=square] table [x=Size, y=Conv1d-MAML, col sep=comma] {csv/output.csv};
    
    \addplot[color=green!50!black] table [x=Size, y=Baseline, col sep=comma] {csv/output.csv};
 
    \addplot[color=blue, mark=triangle*, dashed] table [x=Size, y=MLP-SS, col sep=comma] {csv/output2.csv};
    \addplot[color=blue, mark=*, dashed] table [x=Size, y=GNN-SS, col sep=comma] {csv/output2.csv};
    \addplot[color=blue, mark=square*, dashed] table [x=Size, y=Conv1d-SS, col sep=comma] {csv/output2.csv};
    \addplot[color=orange, mark=triangle*, dashed] table [x=Size, y=MLP-MAML, col sep=comma] {csv/output2.csv};
    \addplot[color=orange, mark=*, dashed] table [x=Size, y=GNN-MAML, col sep=comma] {csv/output2.csv};
    \addplot[color=orange, mark=square*, dashed] table [x=Size, y=Conv1d-MAML, col sep=comma] {csv/output2.csv};

    \nextgroupplot[
    width=3cm,
    height=3cm,
    ymin=0,
    ymax=0.4,
    xmin=0,
    xmax=10,
    legend style={font=\small, draw=white!15!black, at={(1.5,0.5)},anchor=east}
    ]
    
    \addlegendimage{empty legend}
    \addlegendentry{\hspace{-0.5cm}\textit{Learning paradigm}};
    \addlegendimage{color=blue}
    \addlegendentry{SimpleShot};
    
    \addlegendimage{color=orange}
    \addlegendentry{MAML++};
    
    \addlegendimage{color=green!50!black}
    \addlegendentry{Baseline};
    \addlegendimage{empty legend}
    
    \addlegendentry{\hspace{-0.5cm}\textit{Backbone network}};
    \addlegendimage{color=gray, mark=triangle}
    \addlegendentry{MLP};
    \addlegendimage{color=gray, mark=o}
    \addlegendentry{GNN};
    \addlegendimage{color=gray, mark=square}
    \addlegendentry{CNN};
    \addlegendimage{empty legend}
    \addlegendentry{\hspace{-0.5cm}\textit{Hidden layers}};
    \addlegendimage{color=gray}
    \addlegendentry{1};
    \addlegendimage{color=gray, dashed}
    \addlegendentry{2};

\end{groupplot}
\end{tikzpicture}
\vspace{-0.5cm}
\caption{Average accuracy on 500 5-way 5-shot tasks from the validation split as a function of various hyperparameters. These results are used to select the architecture of the best backbone for the evaluation on novel classes. As PT+MAP uses the same backbone as SimpleShot, it is not represented here.}
\label{fig:best_val_acc}
\end{figure}
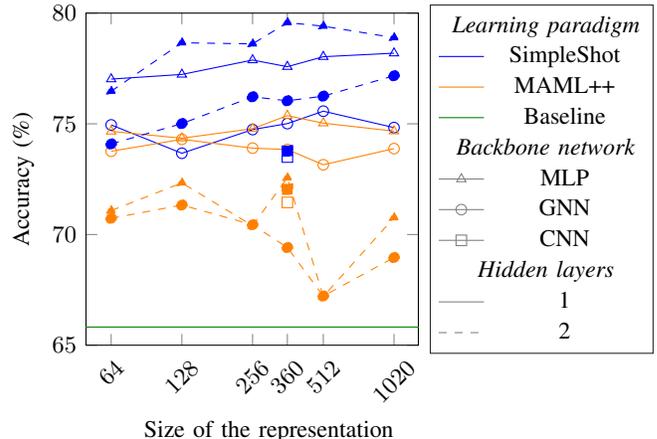

We also look at the influence of the backbones and their hyperparameters on each considered learning paradigm. 
In Table~\ref{tab:best}, the results with GNN are worse than the ones with MLP. GNN fails at better leveraging the underlying structure of this dataset whereas it is naturally supported by a graph. This may be due to the way GNN exploits the graph data structure. Here, GNN diffuses the input on a brain graph. Intuitively, it amounts to averaging the value of the input on each ROI with the values of the regions connected to it on the graph. In some applications, this diffusion acts as a denoising process. Here, the brain signals in some tasks might be strongly localized (e.g. motor tasks) while others engage a distributed set of distant brain regions~\cite{cohen2016segregation}. 

In Fig.~\ref{fig:best_val_acc}, we report the average accuracy obtained on the validation tasks as a function of the size of the representation of the inputs while varying the few-shot paradigms, the backbones and the number of hidden layers. As a reminder, what we call representation of an input is the set of features obtained at the output of the feature extractor. Varying the hyperparameters of each architecture on the validation split enables to select the best hyperparameters for the experiments on the novel classes. As PT+MAP uses the same trained backbone as SimpleShot, it is not represented here. In Fig.~\ref{fig:best_val_acc}, we observe that the number of hidden layers and the size of the representation seem to have little influence on overall performance, even though the trend is towards larger dimensions for SimpleShot and smaller for MAML++.

\subsection{Comparison of learning paradigms for a fixed size of representation}
As shown in the previous experiment, the best representations are not necessarily of the same dimension for various backbones. To provide a fair comparison, we report, in Table~\ref{tab:best} (backbones with representations of size 360), the performance obtained on the novel dataset when all representations are of size 360. Note that we chose 360 because all CNN feature extractors output representations of this dimension. In our case, there are new results only for GNN. We observe that previous conclusions are still valid.

\subsection{Influence of the split}
Finally, we perform experiments with another random split of the classes into training/validation/novel datasets to assess that few-shot learning methods still perform better than the simple baseline. The detailed results are given on our \href{https://github.com/mbonto/fewshot_neuroimaging_classification}{GitHub} repository. We get the same conclusions.

\section{Conclusion}
\label{sec:conclusion}
In this paper, we introduced a neuroimaging benchmark dataset for few-shot learning. By comparing various few-shot methods using several backbone architectures, we achieved engaging results: few-shot vision methods perform well in the context of neuroimaging data. However, the backbone architectures we tested had little influence on the results. In perspective, we will investigate how to conceive a backbone that better leverages the graph data structure of this dataset.

This work opens up a pathway towards the application of deep learning to decode brain signals, where generally few examples are available. We plan to extend this study to 3D time-varying fMRI brain signals. In future work, it would also be interesting to evaluate the few-shot performance over cross-domain datasets.

\bibliographystyle{IEEEtran}
\bibliography{IEEEabrv.bib,refs,20_ML_brain}

\end{document}